\PassOptionsToPackage{table}{xcolor}
\documentclass{article}

    \usepackage[final]{neurips_2024}

\usepackage[utf8]{inputenc} % allow utf-8 input
\usepackage[T1]{fontenc}    % use 8-bit T1 fonts
\usepackage{hyperref}       % hyperlinks
\usepackage{url}            % simple URL typesetting
\usepackage{booktabs}       % professional-quality tables
\usepackage{amsfonts}       % blackboard math symbols
\usepackage{nicefrac}       % compact symbols for 1/2, etc.
\usepackage{microtype}      % microtypography
\usepackage{xcolor}         % colors
\usepackage{enumitem} % Add this to the preamble

\usepackage{booktabs}
\usepackage{graphicx}
\usepackage{multicol}
\usepackage{multirow}
\usepackage{subfigure}
\usepackage{subcaption}
\usepackage{comment}

\newcommand{\ow}[1]{\textcolor{orange}{orion: #1}}
\newcommand{\llq}[1]{\textcolor{blue}{linlu: #1}}

\usepackage{amsmath,amsfonts,bm}

\def\eqref#1{equation~\ref{#1}}
\def\1{\bm{1}}

\DeclareMathAlphabet{\mathsfit}{\encodingdefault}{\sfdefault}{m}{sl}
\SetMathAlphabet{\mathsfit}{bold}{\encodingdefault}{\sfdefault}{bx}{n}

\usepackage[ruled,vlined]{algorithm2e}
\SetArgSty{textup}

\usepackage{hyperref}
\usepackage{cleveref}

\usepackage{url}
\usepackage{xspace}

\newcommand{\method}{\textsc{CoGEX}\xspace}
\newcommand{\search}{\textsc{CoTACS}\xspace}
\newcommand{\searchlong}{\textsc{CoTACS}: \textsc{\textbf{Co}GEX} \textbf{T}ask \textbf{A}daptation via \textbf{C}ode \textbf{S}earch\xspace}
\newcommand{\dataset}{\textsc{CoGEX}\xspace}
\newcommand{\myparagraph}[1]{\noindent\textbf{#1}\hspace{1mm}}

\usepackage{colortbl}

\usepackage{listings}

\lstdefinestyle{base}{
  language=Python,
  emptylines=1,
  breaklines=true,
  commentstyle=\color{gray},
  basicstyle=\scriptsize\ttfamily,
  moredelim=**[is][\color{red}]{@}{@},
}

\def\mystrut(#1,#2){\vrule height #1pt depth #2pt width 0pt}

\definecolor{lightgreen}{RGB}{223,255,219}
\definecolor{lightred}{RGB}{255,219,219}
\definecolor{mygray}{RGB}{220, 220, 220}

\newcommand{\redtext}[1]{\colorbox{lightred}{\mystrut(.4, .4) #1}}
\newcommand{\greentext}[1]{\colorbox{lightgreen}{\mystrut(.4, .4) #1}}
\newcommand{\greytext}[1]{\colorbox{mygray}{\mystrut(.4, .4) #1}}
\newcommand{\up}{\cellcolor{lightgreen}}  
\newcommand{\dn}{\cellcolor{lightred}}  
\newcommand{\same}{\cellcolor{mygray}}  

\title{Learning to Reason via \\ Program Generation, Emulation, and Search}
               
\author{Nathaniel Weir\thanks{Work done in part during internships at Allen Institute for AI.
}$^{\hspace{4pt}\dagger}$ \\
Johns Hopkins University\\
\texttt{nweir@jhu.edu} \\
\And
Muhammad Khalifa$^{*\dagger}$ \\
University of Michigan \\
\texttt{khalifam@umich.edu} \\
\And
Linlu Qiu$^{*}$ \\
MIT \\
\texttt{linluqiu@mit.edu}
\And
Orion Weller$^*$ \\
Johns Hopkins University \\
\texttt{oweller@cs.jhu.edu} \\
\And 
Peter Clark \\
Allen Institute for AI \\
\texttt{peterc@allenai.org}
}

\makeatletter
\def\blfootnote{\xdef\@thefnmark{}\@footnotetext}
\makeatother    
\begin{document}

\maketitle
\begin{abstract}
% Language models (LMs) are generally trained to follow user instructions in order to maximize the amount of tasks they can perform.  
% The standard way of performing this instruction tuning is done by providing the model with a paired training set of natural language instructions and their outputs.
% The standard practice of instruction tuning generally trains language models (LMs) to following natural language instruction by providing natural language outputs. 
% However, a separate line of 
  Program synthesis with language models (LMs) has unlocked a large set of reasoning abilities; code-tuned LMs have proven adept at generating programs that solve a wide variety of algorithmic symbolic manipulation tasks (e.g. word concatenation). However, not all reasoning tasks are easily expressible as code, e.g. tasks involving commonsense reasoning, moral decision-making, and sarcasm understanding. %  are examples of reasoning tasks involving ``softer'' or more fuzzy reasoning that can not be delineated in code. 
%But what about tasks with softer or more fuzzy types of reasoning such as commonsense?
% judgemental steps (e.g., ``determine the most likely location of $\langle\text{object} \rangle$ ''),
% not easily formulated as code? 
% We propose to have the LM generate {\it partial programs}, where ``soft'' leaf function
% calls (e.g., likely
% $\_$location(object)) are left undefined, and then second use the LM to {\it simulate}
% execution of the program, including the likely results of such function calls, to
% produce both intermediate results and a final answer.
% This work challenges the assumption that instruction tuning data should be only formatted in natural language, and proposes to format it using code instead.
% We propose \textbf{Instru}ction Tuning by Emulating \textbf{Code} Execution (\method) which trains LMs to accept a natural language instruction, generate a corresponding Python function code, and emulate the program's execution to produce an output. 
% This raises a question: Can we bring the benefits of reasoning over code to such reasoning tasks?
Our goal is to extend an LM's program synthesis skills to such tasks and evaluate the results via {\it pseudo-programs}, namely Python programs where some leaf function calls are left undefined. To that end, we propose, \textbf{Co}de \textbf{G}eneration and Emulated \textbf{EX}ecution (\method).
%trains LMs to follow arbitrary natural language instructions by first generating a corresponding Python program, which is pseudo-executed to produce an output. 
\method works by (1) training LMs to generate pseudo-programs, (2) teaching them to {\it emulate} their generated program's execution, including those leaf functions, allowing the LM's knowledge to fill in the execution gaps; and (3) using them to search over many programs to find an optimal one.
%\pc{Our method, called} \textbf{Instru}ction Following by Emulating \textbf{Code} Execution (\method), trains LMs to follow {arbitrary NL instructions} by first generating a corresponding Python program that is pseudo-executed by the LM to produce an output. 
%Our method, 
% called \textbf{R}easoning by \textbf{E}mulating Program \textbf{Ex}ecution (\method), 
% We fine-tune LLaMA and Code-LLaMa models on a new dataset to take in as input a natural language instruction and generate Python code then compile/execute to return the output. 
%We create a training set for \method instruction tuning and fine-tune LLaMA/CodeLLaMa models to follow instructions in this manner. 
% This paradigm gives us the opportunity to consider learning a new task as finding a program:
To adapt the \method model to a new task, we introduce a method for performing program search to find a single program whose pseudo-execution yields optimal performance when applied to all the instances of a given dataset.
% We find that models fine-tuned using our approach can then benefit substantially from a task-specific program search: finding a single generalizable program that can be reapplied to the items in a given task dataset.
We show that our approach yields large improvements compared to standard in-context learning approaches on a battery of tasks, both algorithmic and soft reasoning. This result thus demonstrates that code synthesis can be applied to a much broader class of problems than previously considered.\footnote[1]{
Our released dataset, fine-tuned models, and implementation can be found at \url{https://github.com/nweir127/CoGEX}.}
% \pc{These techniques thus allow 
% This leads to substantial performance boosts on tasks in text classification, mathematics, and commonsense QAcompared to typical off-the-shelf usage of Llama family models. 
% We release our \method{} dataset and models for public use.  
% \end{abstract}

\blfootnote{\hspace{-1.5mm}$^\dagger$Co-first authors.}

\end{abstract}

\section{Introduction}

\begin{comment}
{\color{red}

from peter:
 
LMs have proven deft at answering algorithmic questions via code synthesis,
e.g., answering "What is the concatenation of the 3rd letter of every word in <sentence>?" by generating and executing Python code.
But what about tasks that require commonsense or judgemental steps (e.g., ``determine the most likely location of <object>''),
not easily formulated as code?
Our goal is to extend a LM's program synthesis skills to such tasks, and evaluate the results.
Our approach is to first have the LM generate {\it partial programs}, where “soft” leaf function
calls (e.g., likely
$\_$location(object)) are left undefined, and then second use the LM to {\it simulate}
execution of the program, including the likely results of such function calls, to
produce both intermediate results and a final answer.
We use a search method blah blah
Our results show blah blah
This is significant, as it shows how an integration of ideas from partial programming, simulated code execution, and
search can allow programmatic-based solutions to be applied to a much broader class of problems than previously considered.

}
\end{comment}

Recently there have been rapid advances in training language models (LMs) to generate code rather than natural language (NL), following the intuition that code may be more effective than NL for certain tasks, such as those requiring complex calculations,
  iteration, or data structure manipulation\citep{chen2022program,gao-etal-2023-pal}. 
% Additionally, utilizing code interpreters can bring extra benefit since the LM only needs to generate a program that achieves the task \citep{li2023chain}.
Although successful, these works have mostly studied tasks conducive to a programmatic paradigm, such as
symbolic manipulation or algorithmic reasoning, i.e., tasks for which a clear compilable program can be devised.
However, it is unclear how to apply this approach to
``softer'' reasoning tasks such as commonsense and social reasoning, where algorithmic solutions are less obvious \citep{zhang2023exploring}. 

\begin{figure}[t!]
    \centering
        \includegraphics[width=.95\textwidth]{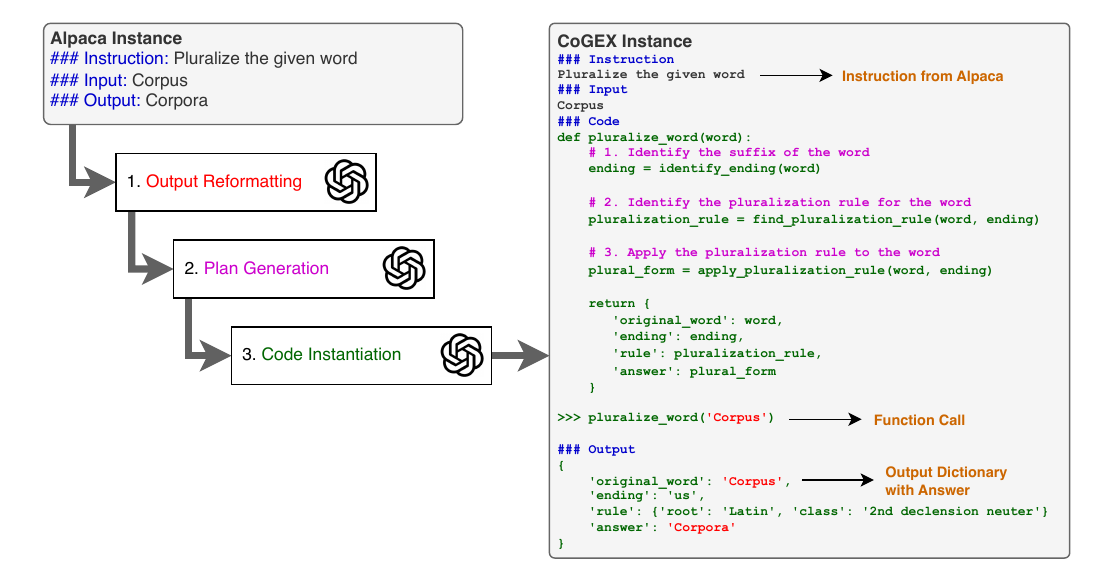}
    %     \quad 
    % \includegraphics[width=0.55\linewidth]{figures/alpaca-code.pdf}
    \caption{Example from the \dataset{} dataset automatically converted from an Alpaca~\citep{taori2023alpaca} instance via LLM prompting. We train the model to receive the instruction and input and generate the Python program and function call (as an intermediate), before outputting the final dictionary that contains the answer and any intermediate reasoning steps. 
    \vspace{-1em}
    }
    \label{fig:data-construction}
\end{figure}

%\llq{should we mention the motivation of using LM as the code emulator? e.g. these approaches mostly rely on using python interpreter. However, for many general tasks, it is less obvious how to write and execute a program?}
%\nw{agreed, we should lean more on that}
Our goal is to expand an LM's program synthesis skills to such softer reasoning tasks. Our approach builds on the insight that, beyond generating code, LMs can also \textit{emulate the execution} of code. This includes handling function calls that are defined only by name and documentation, even if they lack a full implementation. We refer to these types of programs---where only the function skeletons are provided without actual code implementations---as \textit{pseudo-programs}. Such pseudo-programs can encompass both well-defined reasoning steps, such as mathematical or algorithmic operations, as well as function calls representing less precise reasoning, such as commonsense logic. This work investigates whether generating and pseudo-executing such programs can effectively address soft reasoning tasks in addition to traditional algorithmic problems.

To achieve this, we propose a novel approach:
training models to follow NL instructions by generating a program and then \textit{emulating} that program's code execution.
% \footnote{We refer to the generated non-compiled code as \textit{code} or \textit{program} interchangeably.} 
Our paradigm, called \method, 
% trains a model to perform instruction following by flexibly generating and executing its own program in well-formed Python syntax.
% Our method 
changes the inference process to \textbf{(1)} generate a Python function given an arbitrary instruction and optional input,  \textbf{(2)}
generate a call to that function, and \textbf{(3)} produce the result of simulating its execution. 
Unlike other work~\citep{zhang-etal-2023-natural,li2023chain}, we do not use a Python interpreter to execute any code; rather, the model is trained to emulate execution.
%\llq{should we mention \citet{chae2024language, li2023chain}, i guess the main difference is they are still mostly focusing on symbolic tasks but we consider general tasks?}
This allows the generated code to deliberately include calls to underspecified functions (i.e., where only the function name and documentation are included), by allowing the LM emulator to \textit{fill in} the functions implementations using its latent knowledge.
We train the model to not only output the result of pseudo-executing the code but also the results of intermediate function calls in the code.
%This gives us an improved level of interpretability and systematicity.
\method thus suggests a way to leverage the flexible ad-hoc reasoning abilities of LMs while encouraging programmatic reasoning via the program structure.
We train \method models by adapting the recent Alpaca instruction tuning dataset~\citep{taori2023alpaca} into a set of analogous Pythonic examples by prompting GPT-4 to perform the conversion, and then use the resulting \dataset dataset to fine-tune smaller (7B and 13B) LMs to answer instructions via code.

% In addition, we develop an extension of this paradigm, namely:
The \method{} paradigm allows us to explore a new way to learn new tasks:
identifying a {\it general} program that applies across a task, such that new task instances can be solved by
emulating calls to that one program.
Inspired by work on hard prompt tuning~\citep{wen-etal-2024-hard} and example selection for in-context learning~\citep{gupta-etal-2023-coverage}, 
we introduce a search procedure that uses the frozen
\method{} model to try out many programs on a set of training examples and identify which single program is optimal for the dataset.
The procedure, termed \searchlong{}, performs no parameter updates and requires saving nothing more than a program string.
% \ow{We don't require the user to give these programs right?} \nw{right, the model finds the program automatically}

We evaluate over a diverse suite of reasoning tasks, including commonsense QA, text classification, and math datasets. 
These datasets cover symbolic tasks that could conceptually benefit from programmatic operations and standard natural language tasks whose solutions might not be easily described in code.
We find that applying \search{}  leads the \method{} models to substantially outperform the comparable NL-based LM using the same original checkpoint and the same set of training examples available for in-context learning, even in the few-shot regime.
\search{} thus gives us one way to fit a model to a new dataset without having to perform any gradient descent or parameter updates, both for algorithmic and softer reasoning datasets.
% \pc{rather, we} find and save a task-general program. 
% This makes \search a particularly helpful method, even when LM providers do not wish to release model weights.
% We show that \search{} provides these benefits over .
% While existing work on code-based reasoning has either primarily considered symbolic manipulation tasks~\citep{gao-etal-2023-pal} or required additional entailment classifiers~\citep{zhang-etal-2023-natural}, ours requires only the LM checkpoint and applies broadly to general NLP and reasoning tasks.
% and a systematically searched-for, task-specific program.
% \llq{do we want to say \method as a framework? or a model? or a dataset? naming seems not very consistent}
Our contributions are thus:
\begin{enumerate}
\vspace{-2mm}
\item A novel reasoning paradigm, \method{}, that trains language models to 
  generate and emulate the execution of pseudo-programs. 
  \method is a general paradigm that allows LMs to leverage code for different types of reasoning. %\llq{do we still want to claim it's a instruction tuning paradigm, or just fine-tuning?}
\item A program search method, \search, enabling a task-general program suitable for a dataset (rather than a single instance) to be found using a \method model.
\item A dataset, derived from the Alpaca instruction tuning dataset, for training \method{} models.
% \item Experiments demonstrating the effectiveness of \method{} and \search{}.
\end{enumerate}

Overall, this work provides a significant step in showing how code generation can be applied to a much broader class of problems than previously considered.
%   The \dataset{} dataset, fine-tuned models, and code for the \search{} program search algorithm can be found at \url{https://www.booking.yeah}.}

% \vspace{-1mm}

\section{Approach}

% \vspace{-1.1mm}
\label{sec:approach}
In this section, we start by formalizing our approach and describing our data construction process (\S\ref{sec:data_recast}). 
We then describe our program search approach to tune a \method model on a given task through program search (\S\ref{sec:code_search}). 
%This approach uses only the \method{} model's generations and avoids the use of any stochastic gradient descent or external LMs.
% \vspace{-1.1mm}

% \begin{figure}
%     \centering

%     \caption{Caption}
%     \label{fig:enter-label}
% \end{figure}

\subsection{Method: \method}
\myparagraph{Formulation.}
Our goal is for the model to execute a given task by simulating code execution. That means our model will take as input the task description, generate a Python program, and simulate the expected output of executing that program.  
Formally, given a natural language (NL) task description $I$, optional input argument $A$, Python function $F$, function call $C$, and output dictionary $O$ designating the output from the program pseudo-execution, the LM will take $\langle I, A\rangle$ as input and generates $\langle F, C, O\rangle$ as output. Since the process is sequential, \method models can work as either a reasoner $f_\text{reasoner}(I, A) \rightarrow (P, C) \rightarrow O$ 
% \llq{is it supposed to be (f(I, A) $\rightarrow$ (P, C)?}
%that instantiates an instruction-specific Python function, function call, and corresponding execution result, 
or as a call-instantiating and execution-emulating model $f_\text{emulator}(I, A, P) \rightarrow C \rightarrow O$  that takes a pre-specified program $P$ and applies it to the variable arguments $A$. This latter formulation enables searching over the space of task-specific programs: searching for one $P_\text{\text{task}}$ to solve a class of problem (e.g., emotion classification) and then applying $f_\text{emulator}(I, A_i, P_\text{\text{task}})$ to emulate its execution on each instance $A_i$ of that problem. We expand on program search in \S\ref{sec:code_search}.

\myparagraph{Training Data Construction.}
\label{sec:data_recast}
As we want a general-purpose dataset that spans tasks with diverse reasoning requirements, we choose the Alpaca instruction tuning dataset \citep{taori2023alpaca}. Following \citet{peng-etal-2023-instruction}, %who use an LLM to construct data for fine-tuning smaller models, 
we rely on GPT-4\footnote{The dataset was constructed between August 7th--26th, 2023 using the \texttt{gpt-4} model in the OpenAI API.} to convert the Alpaca dataset into their \method versions. Specifically, every NL instance in the Alpaca dataset is mapped into a corresponding \method version. 
% A naive approach is to directly prompt GPT to convert a given Alpaca instance to the desired code format. However, our initial trials showed that doing the conversion in a single pass yielded incorrectly formatted outputs from GPT-4. Therefore, w
We split the conversion process into three steps, each of which involves prompting GPT-4 with the output from the previous. This stepwise approach proved more effective than directly prompting GPT to convert each instance to code in one shot.

As depicted in \Cref{fig:data-construction}, the three steps are: 
\textbf{(1)} converting the outputs and (optional) inputs into Pythonic data structures like strings, lists, and integers whenever relevant as determined by GPT-4;
\textbf{(2)} generating an instruction-specific plan, or a series of NL steps that should perform the task for any potential input;
\textbf{(3)} instantiating the plan as a Python program whose inline comments are the plan steps and whose output is a dictionary containing all intermediate and final outputs that the LM believes would result from executing each step. Prompts for all steps can be found in \Cref{app:prompts}.

Importantly, we allow for GPT to include \textit{undefined} functions, e.g., \texttt{identify\_ending()} and \texttt{find\_pluralization\_rule()} in \autoref{fig:data-construction}. The goal is to leverage the LM knowledge to fill in the semantics of these undefined functions when emulating the execution of a given program. 
%This allows the model to e.g. apply a function to a paragraph of text without worrying about applying any typical NLP preprocessing operations.
%Since we cue the model to emulate the full execution of the program end-to-end, these under-specified module calls give the model the flexibility to perform ad-hoc reasoning when appropriate in an otherwise quasi-symbolic series of operations.
In addition, we include the program's \textit{intermediate results} in the output dictionary before the final answer to encourage the model to stick to the NL reasoning plan delineated in the program comments. %in addition to improving the interpretability. 
%actually use these program lines for reasoning about the task, as opposed to foregoing the program and reasoning by some other means.\llq{should we also mention this also improves interpretability?} \mk{I suppose we need an ablation for this.}
After defining the program, we cue GPT-4 to \textit{call} the function on an argument, e.g. \texttt{pluralize\_word(`corpus')} which can reflect the optional Alpaca example input, or can reflect specific details from the instruction itself. Our prompts encourage GPT-4 to write a program that is as general purpose as possible and not tied to a specific input: e.g. \texttt{pluralize\_word(word)} is preferable to \texttt{pluralize\_corpus()}.

% \subsubsection{Model Training}
Fine-tuning any LM on the resulting \method{} dataset creates our desired model, which accepts any task description/input combination and responds by dynamically generating a Python program and then emulating its execution.

% \vspace{-1.5mm}
\subsection{Program Search: \search}
\label{sec:code_search}
A \method model can generate a new program for any new task instruction and instance; 
however, some programs might be more or less effective at performing the task.
How can we find the optimal program for a specific task, especially when some training data is available? As \method relies on argument-accepting pseudo-programs, it naturally enables program optimization. Given multiple examples of a new task, we can search for \textit{one} program that performs well, and then apply the same program to new test examples by invoking the program with different input arguments.
%This suggests that we can search for an optimal program that is generalizable across examples within the same task.\footnote{Although not exactly comparable, this is similar to using a best set of examples for in-context learning.}
% This enables a task adaptation method that uses code search to identity the optimal programs that generalizable across multiple examples. 
% When we have multiple examples for a new task, we can simply 
% Since programs allow take different input arguments for different function calls
% is it more effective to generate a separate program for every instance, or find one program that can be reapplied to every datapoint?
% We introduce a code search optimization procedure that seeks to answer this question. 

Our search process, \searchlong{}, finds a single program that optimizes a \method{} model for a particular task dataset, enabling adapting a \method model to a given task without learning any weight parameters. %Rather than updating the model checkpoint via gradient descent,
We learn a new dataset simply by using a finetuned \method model to generate and then evaluate many program candidates to find the one that best fits the given dataset.
As described in \autoref{alg:search}, we split a dataset $D$ of argument and output pairs $(a_i, o_i)$ into a small training set ($n$ $=$ $300$ in experiments) and a larger development set; we then generate a separate code candidate for every training item and retain the programs with decent performance on the training set.\footnote{To ensure quality of the retained programs, a user-defined threshold $\alpha$ is used to keep only the programs whose training performance is at least $\alpha$. If no program achieves $\alpha$ after a fixed number of trials, we use the best performing sampled so far.} We then rank these programs according by their performance on the development set. For certain tasks, we find it beneficial to find \textit{multiple} programs for a task and then at test time take a majority vote across the \method{} model's responses using each code. 
To accomplish this, we retain some top-$k$ performing codes over the development set.

\begin{algorithm}[h!]
{\small
\SetAlgoLined
\LinesNumbered
\SetKwInput{KwInput}{Input}

\KwInput{\method{} model \(f\), Dataset \( D = \{(a_1, o_1), (a_2, o_2), \ldots\} \), Instruction \( I \), number of code candidates \( n \), minimum training performance \( \alpha\), task metric \( \delta\)}
\KwResult{Optimal programs \( P_D \) that maximize model performance on \( D \)}
\BlankLine
% Initialize model with instruction \( I \)\;
\( \text{Programs} \leftarrow \emptyset \)\;
\( \text{TrainSet} \leftarrow \text{RandomSample($D$, $n$)} \)\tcp*{Sample from $D$}\ 
\( \text{DevSet} \leftarrow D \setminus \text{TrainSet}\) \tcp*{Remaining \( |D| - n \) examples serve as dev set}
\BlankLine
\For{\( (a_i, o_i) \) in \( \text{TrainSet} \)}{
    \( p_i \leftarrow f(I, a_i) \)\tcp*{Sample a program for the instance}\ 
    \( \text{TrainPerf} \leftarrow \texttt{Evaluate}(p_i, \text{TrainSet})\)\;
    \While{\text{TrainPerf} $< \alpha$}{
        \( p_i \leftarrow f(I, a_i) \) \tcp*{Resample code if low performance}
        \( \text{TrainPerf} \leftarrow \texttt{Evaluate}(p_i, \text{TrainSet}) \)\;
    }
    Add \( p_i \) to \( \text{Programs} \)\;
}
\BlankLine
\( P_D \leftarrow \text{argmax}_{P = \{p_1, \dots p_k\} \subseteq \text{Programs}}\sum_{i=1}^k \texttt{Evaluate}(p_i, \text{DevSet}) \)\;
\Return \( P_D \)\;

\BlankLine
\BlankLine

\SetKwFunction{FEvaluate}{Evaluate}
\SetKwProg{Fn}{Function}{:}{}
\Fn{\FEvaluate{$p$, $D$}}{
    \For{\( (a_i, o_i) \) in \( D \)}{
        \( (c_i, \hat{o_i}) \leftarrow f(I, a_i, p) \)\tcp*{Run the model with program $p$}
    }
    \Return \( \frac{1}{|D|} \sum_{i=1}^{|D|} \delta(\hat{o_i}, o_i) \)\tcp*{Average task metric (e.g., exact match)}
    
}
}
\caption{\search search that identifies a set of $k$ programs $P_D$ that best adapts a \method model to new dataset $D$ 
}
\label{alg:search}
\end{algorithm}
% \vspace{-0.25em}

% \llq{i think we could probably better motivate why we want to do learning as code search, so far it's less obvious to me why we transit to this part. Maybe we could say something like: \method encourages programs that are generally applicable across many arguments, which is a desirable property when adapting models to a specific task. Here we extend \method to the task adaptation setting where all examples share the same general program, and this motivates us to perform search to identify the optimal programs.}
% \ow{Always 300, or should we say $N>=1$?} \nw{the curves figure shows when $N \leq 300$}
% \ow{In the algorithm, what is o?}
% \nw{o is outputs}

% \vspace{-.4cm}

\begin{table}[ht!]
\small
\setlength{\tabcolsep}{3pt} 
    \centering
    
\caption{Benchmark results by  \method{} models optimized for each dataset using the \search{} method, compared to the corresponding off-the-shelf Llama-2 checkpoint 
 performing 2-shot reasoning using a BM25 retrieval index of 1000 exemplars. Results are also compared to a zero-shot Alpaca model fine-tuned from the same checkpoint.
The top score per size is \textbf{bolded}.
 Colored cells indicate changes (\greentext{gains}, \redtext{losses}, or the \greytext{same}) relative to the best-performing non-\method{} baseline (Alpaca or 2-shot). 
 Results show that \method{} with \search{} outperforms the baselines on nearly every task and often does so even with only 10 examples. \vspace{3pt}
}
\scalebox{0.9}{
% \begin{tabular}{p{2.2cm}lrrrrrrrrr|c}
\begin{tabular}{p{2.2cm}lcccccccccc|c}
\toprule
& & & \multicolumn{3}{c}{\textbf{Classification}} & \multicolumn{2}{c}{\textbf{Symbolic}} & \multicolumn{2}{c}{\textbf{Math}} & \multicolumn{2}{c}{\textbf{Commonsense}} & 
%  \multicolumn{1}{c}{\textbf{Average}} 
\\
\cmidrule(lr){4-6} \cmidrule(lr){7-8} \cmidrule(lr){9-10} \cmidrule(lr){11-12} 
% \cmidrule(lr){12-12}
& &  $N_{\text{train}}$ & CoLA & Emotn & SST & Coin & WSort & Sum & SVAMP & CSQA & SIQA & \textbf{Avg} \\
\midrule
Alpaca 7B & 0-shot & 0 & 70.6 & 53.4 & 87.3 & 49.5 & 40.0 & 21.6 & 25.7 & 46.8 & 54.1 & 49.9 \\
Llama-2 7B & 2-S BM25 & 1000 & 57.5 & 55.2 & 82.1 & 32.4 & \textbf{45.5} & 35.2 & 34.7 & 45.7 & 46.0 & 48.3 \\
\midrule
\multirow[c]{3}{2.2cm}{\textbf{\method} Llama-2 7B} & \search{} $k=1$ & 10 &  \up 75.0 & \dn 52.2 & \up 86.9 & \up 50.8 & \dn 40.6 & \up 61.3 & \dn 33.3 & \dn 42.3 & \up 50.1 &   56.7 \\
 & \search{} ${k=1}$ & 1000 & \up 78.5 & \up \textbf{56.2} & \up \textbf{91.2} & \up 60.0 & \dn 39.5 & \up 62.8 & \up 41.3 & \up \textbf{52.7} &\up  57.7 & 60.0 \\
 & \search{} ${k=3}$ & 1000 & \up \textbf{79.2} & \up 56.0 & \up 90.9 & \textbf{61.9} \up & \dn 40.8 & \up \textbf{63.6} & \up \textbf{42.7} & 52.4 \up & \up \textbf{59.3} & \textbf{60.8} \\
\midrule
Alpaca 13B & 0-shot & 0 & 74.6 & 53.0 & 86.0 & 63.8 & 50.3 & 44.5 & 37.3 & 62.2 & 63.4 & 59.5 \\
\multirow[t]{2}{2.2cm}{Llama-2 13B} & 2-S BM25 & 1000 & 78.9 & 54.0 & \textbf{92.4} & 48.6 & 50.0 & 38.7 & 46.0 & 63.4 & 62.8 & 59.4 \\
\midrule
\multirow[c]{3}{2.2cm}{\textbf{\method} Llama-2 13B} & \search{} $k=1$ & 10 & \up 80.9 & \up 55.1 & \dn 88.9 & \dn 58.0 & \up 51.3 & \up 61.3 & \dn 42.8 & \dn 59.2 & \dn 57.6 & 61.7 \\
& \search{} ${k=1}$ & 1000 & \up 81.0 & \up  56.4 & \dn 92.1 & \up 65.7 & \up  50.8 & \up  \textbf{64.0} & \up 48.3 & \same 63.4 &  \up 64.6 & 65.1 \\
& \search{} ${k=3}$ & 1000 & \up \textbf{81.0} & \up \textbf{56.6} & \dn 92.1 & \up \textbf{69.5} & \up \textbf{51.6} & \up 63.9 & \up \textbf{50.3} & \up \textbf{64.0} & \up \textbf{65.5} & \textbf{66.1} \\
\bottomrule
\end{tabular}
}
% \vspace{-1em}
\label{tab:main_results}
\end{table}

\section{Experiments and Results}
Below we describe the training and evaluation of \method{} models. We first show overall performance across a wide array of tasks compared to off-the-shelf baselines (\S\ref{sec:main-results}) then ask a series of follow-up research questions investigating ablated scenarios (\S\ref{sec:ablations}).

\subsection{Experimental Setup}
\label{sec:exp-setup}
\paragraph{Model Training.}
% As \method models, 
% we fine-tune  and CodeLlama \citep{roziere2023code}. 
We fine-tune the 7B and 13B variants of  Llama-2 \citep{touvron2023llama}. We use parameter-efficient training via Low-rank adaptation (LoRA) \citep{hu2021lora} with a rank of $r=16$, dropout rate of $0.05$, and LoRA weights added to the query, key, value, and output matrices in all layers. We train all models for five epochs using a batch size of $32$ and a learning rate of $0.0003$. As a validation set, we randomly sample 2K examples from the training set and keep the checkpoint with the lowest perplexity on the validation set for testing. Model training was done on a server with 128GB of RAM and 2 Nvidia A6000 48GB GPUs. On our dataset, training a single 7B model and 13B models took around 12 and 20 hours, respectively. To ensure a fair comparison, all the baselines are trained with the exact same hyperparameters.

\myparagraph{Datasets.}
We measure \method model performance on a variety of benchmarks ranging from symbolic manipulation to commonsense and social reasoning. 
We choose these datasets as representatives of tasks that involve complex reasoning and tasks whose solutions cannot be easily described in code.
%This lets us explore whether our code-based reasoning models are better equipped for the kinds of tasks that conceptually align with programmatic operations versus typical natural language reasoning tasks. 
As for symbolic and math reasoning, we use the Word Sorting task from BIG-bench hard~\citep{srivastava-etal-2022-beyond, suzgun-etal-2022-challenging}, the math word problem dataset SVAMP~\citep{patel-etal-2021-nlp}, 
the coin flip tracking dataset from \cite{wei-etal-2022-chain}, and the large number arithmetic task (referred to as {Sum}) from \cite{zelikman-etal-2022-star}. 
For the last, we use the 5-digit examples for training and 6-digit for testing. We measure string-normalized exact match accuracy for all tasks. Following \cite{zhang-etal-2023-natural}, we evaluate on a series of \textbf{Text Classification} datasets: CoLA (2 labels)~\citep{warstadt-etal-2019-neural}, Emotion (6 labels)~\citep{saravia-etal-2018-carer}, and SST2 (2 labels)~\citep{socher-etal-2013-recursive}.
We also evaluate on the \textbf{Commonsense Reasoning} datasets CommonsenseQA~\citep{talmor-etal-2019-commonsenseqa} and Social IQa~\citep{sap-etal-2019-social}, which are 4- and 5-way multiple-choice datasets. We hand-write the instruction $I_{\text{task}}$ for these datasets as they do not provide any.

\myparagraph{Code Search.} For all datasets, we use a maximum of 1000 training examples. %\footnote{CoinFlip has only 245 training examples.}
We use $n=300$ training items to generate candidate codes and evaluate them on the remaining 700 items to identify the most generalizable ones. 
We experiment with retaining the top-$k \in \{1,3\}$ performing programs for use at test time. For $k=3$, we take a majority vote of answers, breaking ties randomly.
We use sampling temperature $t=0.7$ when generating candidate codes and $t=0.05$ to generate answers given a program and argument.
We report results on the released dev sets of all considered tasks.

\myparagraph{Baselines.} 
We consider two baselines that represent standard practices for adapting an LM to a new task: (1) few-shot prompting using the off-the-shelf Llama-2 and CodeLlama models and (2) zero-shot prompting using the Llama-2 models instruction-tuned from the original Alpaca dataset.\footnote{We also experimented with few-shot prompting the Alpaca models, but found them to perform significantly worse than zero-shot, which is likely a side effect of instruction tuning.} For the in-context learning baseline for (1) we use the same 1000 training data-points as \search{} and optimize the examples by retrieving the most similar few-shot examples using BM25. For zero-shot alpaca models, we use the standard Alpaca-7B and -13B models. 
% \llq{nate, i changed this, but please feel free to edit. also maybe we can add self-consistency baseline here? and maybe results on 0-shot cot?}
% \nw{I'm not sure whether to include the figure 4 baselines here or in the ablation studies section}
% \nw{re self consistency: it really didn't effect much to add self consistency. plus or minus a point here and there. I think we can comment as much without showing the full results for it.} \llq{sg}
% To illustrate the practical effectiveness of the \method{} models, we compare them to off-the-shelf LLama-2 and CodeLlama checkpoints given the same training data.
% We constrain both scenarios to not use any gradient descent:
% we use \search{} to tune the \method{} models on 1000 training datapoints,
% and hook up the off-the-shelf baselines to a 1000-item BM25 retrieval index that at test time constructs a few-shot model prompt.
% As 1000 examples is relatively large for few-shot prompting, we also show the performance when using \search{} with just 10 training examples, again comparing to in-context learning.
As \search{} might not require many training examples to achieve strong performance, we compare the 1000-example \search{} run with one that only uses 10 total examples to generate and evaluate candidates.

% \llq{do we have this result?} \nw{yeah it's line 2 of each section} \llq{is it already part of figure 2? if so maybe we can only focus on cases where we have 1k examples here? also is it supposed to be 1000 instead of 100?}

% We also compare against Alpaca 7B and 13B instruction-tuned from Llama-2 using the Alpaca-Lora codebase.

\subsection{Results}
\label{sec:main-results}
Our main results depicting the difference in performance between \method{} models tuned via \search{} versus off-the-shelf few-shot baselines and Alpaca models are shown in \autoref{tab:main_results}. 
The \search{} method with 1000 training examples outperforms the baselines for a large majority of tasks and models (8/9 tasks for both Llama-2 7B and 13B).
\method{} shows particularly strong gains over the baselines in the Sum and coin flip tracking tasks (+10-20\%) as expected due to its code-related nature.
We observe that the \search{} method with $N_{\text{train}} = 1000$ training examples performs best on average across the 9 tasks, and still performs better than the baselines with only $N_{\text{train}} = 10$ examples.
Retaining the top $k=3$ programs instead of 1 improves performance in most cases (+1\% average).
% , with the exception of CodeLLama 13B $k=1$ and $k=3$ perform about the same.

\myparagraph{Instruction Following.} As our models are trained on instruction following in code, can they still perform instruction-related tasks as well as models trained on text-only Alpaca? We verify this by using \texttt{alpaca-eval} to compare Alpaca-7B against our \method-7B model trained from the same base Llama model. We find a similar win rate (50\% within the 2 SD range) indicating similar instruction following ability. Thus we can see that training on code-based instructions does not hurt standard instruction-following abilities, while opening up many possibilities for program search.

\begin{table}[t!]
    \centering
    \caption{Difference in performance by \search{} $k=3$ comparing Llama-2 vs Code Llama \method{} models (`+' implies Llama-2 better). We see that Code Llama is more effective for some tasks but worse on others, while the 13B version performs worse than the 13B Llama-2 on all but 2 tasks.
    \vspace{3pt}
    }
    
    \setlength{\tabcolsep}{3pt} 
\scalebox{0.9}{
    \begin{tabular}{c|ccccccccc}
    \toprule 
   Model Size  & CoLA & Emotn & SST & Coin & WSort & Sum & SVAMP & CSQA & SIQA \\ \midrule
     7B    &  \up +1.3 & \dn  -0.8 & \up +1.2 & \dn -10.5 & \dn -4.5 & \up +6.2 & \up +2.7 & \dn -3.6 & \up +4.2\\
     13B    & \up +0.5 & \up +0.2 & \up +0.9 & \dn -12.4 & \up +2.4 & \dn -9.3  & \up +3.0 &\up  +6.8 & \up +4.7 \\ \bottomrule
    \end{tabular}
    }
    \vspace{-2mm}
    \label{tab:code_llama_delta}
\end{table}

\myparagraph{Effect of Code Pre-training.}
As we are fine-tuning LMs on code data and then evaluating them on tasks that are more or less code-related, a natural question to ask is whether LMs pre-trained on code datasets yield stronger \method{} models. We investigate this by fine-tuning Code Llama~\citep{roziere2023code} on the \dataset{} dataset instead of Llama.
\autoref{tab:code_llama_delta} shows the resulting change in performance using \search{} ($k$$=$$3$) program search. The Llama-2 models show improved performance on Social IQa (+4\%) but much worse on coin flip tracking (-10-12\%). 
These results do not provide conclusive evidence that Llama-2 models are better or worse than Code Llama on particular task categories.

\subsection{Ablation Studies}
\label{sec:ablations}
\begin{figure}
    \centering
    \includegraphics[width=\textwidth]{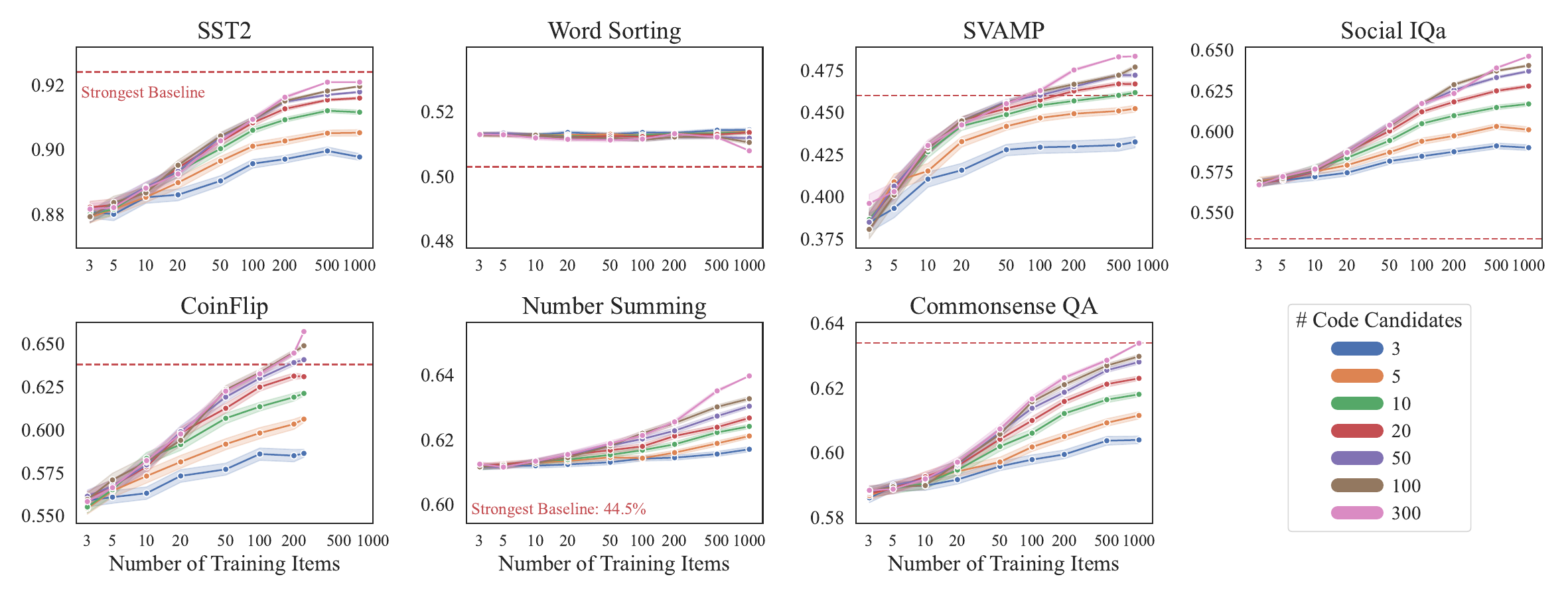}
    \caption{
    Change in \search{} performance using Llama-2 13B as we increase the number of training items from 5 to 1000 and program candidates from 3 to 300. Results are averaged across 1000 trials.
    }
    \label{fig:performance_curves}
\end{figure}

Here we present a series of ablation studies to ask the following questions:

\myparagraph{How many training examples are needed for search?} 
In the above experiments, we chose 1000 training examples and 300 program candidates for the \search{} algorithm. 
This raises the question: how many examples are required to yield the strong performance provided by the search?
We investigate this by simulating the algorithm and sampling 1000 trials with varying numbers of training examples and program candidates. 
% We cap the number of programs by the number of training examples. 
Results are shown in \autoref{fig:performance_curves}. In nearly all cases, performance with 50 or 200 training examples is within a couple of points of the full performance with the 300/1000 configuration.
The performance when sampling 10 code candidates (green) lands within 2 points of sampling 300 candidates on 5 of the 7 datasets.
Benefits do not appear on Word Sorting, as performance lands between 0.51 and 0.515 regardless of configuration. This suggests that the range of quality in generated programs for the Word Sorting dataset is much smaller than the others, so picking between just a few candidates is sufficient. Overall, we see that we can significantly reduce the search space and still see large gains on most tasks.

\begin{figure}
    \centering
    \includegraphics[width=\textwidth]{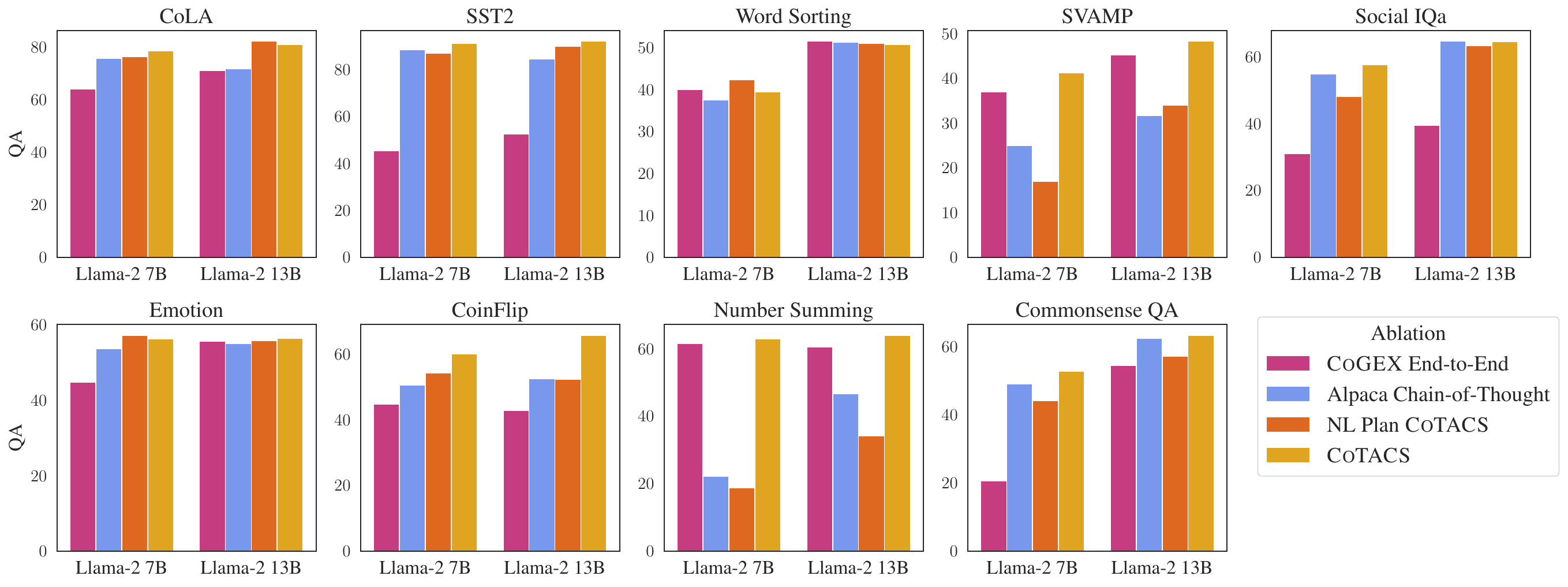}
    \caption{Performance comparison between \search{} ($k$$=$$1$) and various ablations: (1) \method End-to-End that generates separate programs for each instance, (2) chain-of-thought prompting, and (3) searching for an optimal NL plan instead of code program. \search{} consistently equals or outperforms all ablations on all tasks, while each ablation drops in performance on at least 2-3 tasks. 
    % \vspace{-0.5em}
    }
    \label{fig:ablations}
\end{figure}

\myparagraph{Is it better to execute an NL plan instead of Python code?}
We have proposed a mechanism to generate Python programs whose steps are meant to reflect the reasoning process, stated in NL, that answers a given instruction. 
Is the code necessary, or can a model be trained to generate only the plan and achieve the same performance?
We fine-tune a Llama model on a version of the \method{} 52k-item training set where each Python program has been replaced with just the NL steps (removing step 3 of \autoref{fig:data-construction}). This \textbf{NL Plan} model still returns the same output dictionary with intermediate results.
To fit the plan-only model to a dataset, we run the \search{} algorithm but sample and retain the NL plans instead of programs. We see in \autoref{fig:ablations} (orange vs gold) that NL Plan \search{} can match the performance of regular program \search{} on some tasks (CoLA, SST, Word Sorting, Emotion), but performs much worse on others, particularly Coin Flip, SVAMP, and Sum. This follows the intuition that these tasks benefit from a programmatic reasoning paradigm.

\begin{figure}[h]
    \centering
    \includegraphics[width=\textwidth]{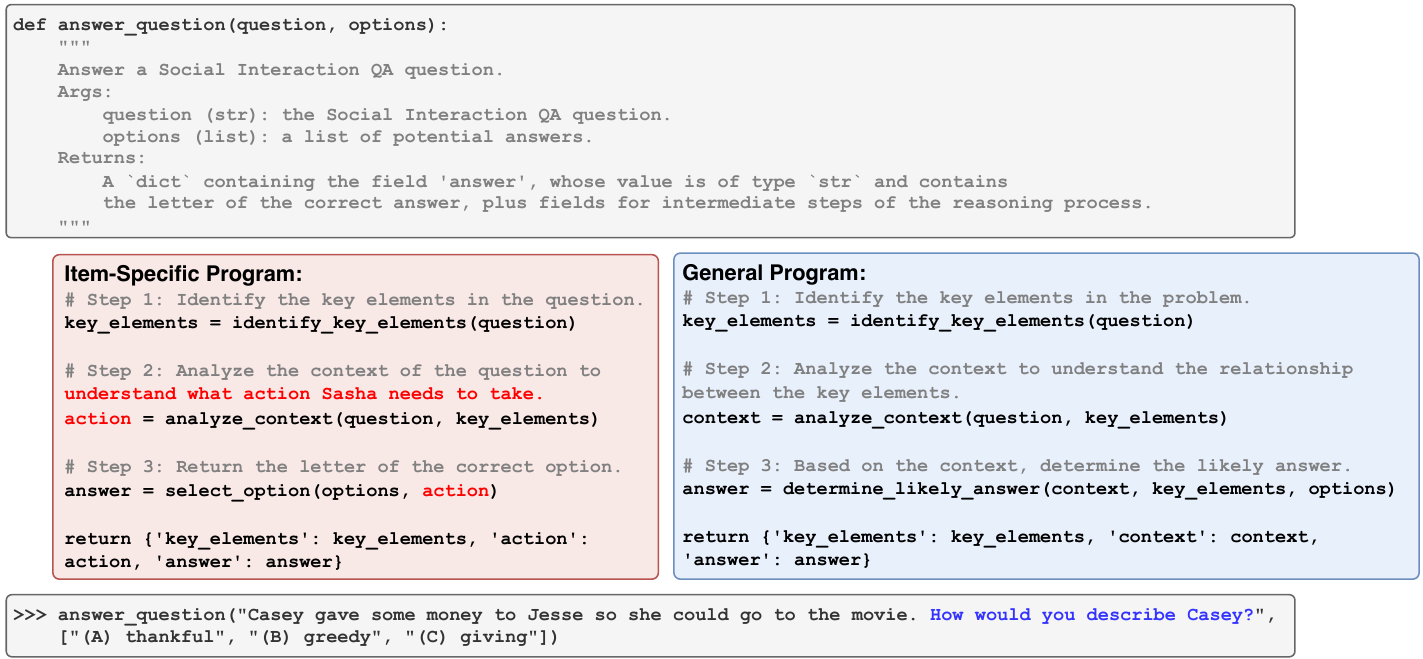}
    \caption{Example \method{} model-generated programs for Social IQa~\citep{sap-etal-2019-social} questions.
    % : one that does not generalize well to the dataset, and one that does. 
    The \textcolor{red}{left item} fits well to a specific SocialIQa question pertaining to question-specific entities but does not generalize well to the dataset, while the \textcolor{blue}{right item} applies more generally to cases such as the instance shown at the bottom, which does not pertain to character actions.
    % Allowing an \method{} model to generate a separate program for each instance in a dataset can yield programs that resemble the left, while 
    Applying \search{} to identify a single program such as the right one shows to improve overall task accuracy. 
    % \mk{maybe it's worth showing the entire prompt (instruction,code, and function call just to make it clear how \method is aplied to specific tasks}
    }
    \label{fig:direct_vs_general}
\end{figure}

\myparagraph{Is it better to find one program or generate a new one for each instance?}
\search{} finds one or multiple programs that can be reapplied to all task instances in a dataset to achieve high performance.
Is this better than letting the \method{} model generate a separate program for every instance?
It might be the case that the latter allows for catering the program to the specifics of a particular task instance-- e.g. in \autoref{fig:direct_vs_general}, where the left (red) \method{}-generated program has steps specifically crafted to identify actions to be taken by a particular person. Finding a single program disallows this flexibility.
We investigate this question by running the model end-to-end on each instance. The \search{} model performs the mapping $f(I_{\text{task}}, A_i, P_{\search{}}) \rightarrow C_i \rightarrow  O_i$ for each task instance $A_i$,
while the end-to-end model performs $f(I_{\text{task}}, A_i) \rightarrow (P_i, C_i) \rightarrow O_i$.
We sample from the latter using temperature $t=0.05$.\footnote{ Increasing $t$ and/or using self-consistency~\citep{wang2023selfconsistency} did not meaningfully affect performance.} 
Results are shown in \autoref{fig:ablations} (maroon vs gold); end-to-end performance is comparable to \search{} only on Word Sorting and Sum. In all other cases, it is substantially worse.

\break % to avoid weird issue when citation is at page end
\myparagraph{Is \search better than chain-of-thought?}
A common practice to elicit systematic reasoning from LMs is to prompt it for the reasoning via some version of ``explain your answer step-by-step''~\citep{kojima-etal-2022-large}. How does this compare to \method{} models on a given dataset? 
We compare \method to zero-shot CoT by prompting our Alpaca models with a task-specific instruction, while additionally appending the instruction to ``think step-by-step'' before producing the answer. \autoref{fig:ablations} (blue vs gold) shows that CoT prompting performs similarly to the NL plan search method; it can approach \search{} performance on some NLP classification tasks but performs worse over SVAMP, Number Summing, and CoinFlip. 

\myparagraph{When is \search{} better than fine-tuning?}
Fine-tuning is a standard practice to adapt an LM to a new task. However, fine-tuning often requires a large amount of training data and storing a new model for each task. Here, we study the impacts of the number of examples on fine-tuning and \search{}. We find that when there are many examples available, fine-tuning achieves stronger performance. However, \search{} is generally better until there are a large number of examples available: it outperforms fine-tuning on 4/9 tasks with 500 examples. This suggests that \search{} can be a lightweight alternative in the low-to-medium shot setup.
% We would expect that fine-tuning a model would be more effective than \search{} when there are many examples and one has the capability to fine-tune and store a new model. 
% In \autoref{fig:finetune} we show that this is true by fine-tuning the same base model on $N$ examples from each task. We see that \search{} is generally better until the there are a large number (500+) of examples available for fine-tuning and even then is a lightweight alternative.

\begin{comment}
\ow{Should we move 3.4.1 to this section and reference the figure in each sub-bullet?}

\subsubsection{Ablation Results}
Results comparing \search{} ($k=1$) to the above described ablations and chain-of-thought are shown in \Cref{fig:ablations}. Across the 9 tasks, \textbf{the regular \search{} method (gold) consistently exceeds or equals all the other methods, while each other method falls short of the others for at least one task}. End-to-End decoding (maroon) is worse in all cases except word sorting. Chain-of-Thought prompting (blue) and NL plan search (orange) perform far worse on coin flip tracking and the math datasets, Sum and SVAMP.
These results illustrate the general applicability of searching over task-specific codes.

\end{comment}

\begin{comment}
\begin{table}[t!]
    \centering
\scalebox{0.9}{
    \begin{tabular}{llccccccccc}
\toprule
 $N_{\text{train}}$ &  &  CoLA & Emotn & SST & Coin & WSort & Sum & SVAMP & CSQA & SIQA \\
\midrule
% & 2-S BM25 & 63.4 & 52.2 & 85.2 & 21.9 & 42.1 & 35.2 & 31.7 & 47.9 & 47.0 \\
& 2-S BM25 & 81.9 & 51.1 & 94.3 & 48.6 & 45.0 & 41.7 & 40.7 & 64.0 & 63.1 \\ 
\multirow[t]{2}{*}{100} & Fine-tuning & 65.8 & 42.5 & 62.7 & 56.2 & 51.6 & 31.1 & 16.3 & 47.3 & 47.9 \\ 
 % & \search{} & \up 80.3 & \up 56.3 & \up 
 % 90.1 &\up  75.5 & \dn 49.4 & \up 72.2 & \up  46.4 & \up  55.1 & \up  57.3 \\
  & \search{} 
& 81.2 & 55.5 & 91.0 & 63.3 & 51.1 & 62.1 & 46.3 & 61.6 & 61.7 \\ 
\midrule
& 2-S BM25 & 81.0 & & 92.7 & 48.6 & 53.4 & & 46.0 & 63.3 &  \\
\multirow[t]{2}{*}{500} & Fine-tuning & 72.6 & 61.9 & 95.0 & 0.0 & 55.5 & 76.0 & 45.3 & 60.4 & 56.3 \\
 & \search{} & 81.5 & 55.8 & 92.1 & 65.7 & 51.3 & 63.5 & 48.3 & 62.8 & 63.8 \\
\midrule
& 2-S BM25 & 78.9 & 54.0 & \textbf{92.4} & 48.6 & 50.0 & 38.7 & 46.0 & 63.4 & 62.8 \\
\multirow[t]{2}{*}{1000} & Fine-tuning & 73.2 & 81.8 & 94.0 & 56.2 & 57.9 & 77.3 & 53.7 & 74.5 & 69.4 \\
 & \search{} & 81.0 & 56.4 & 92.1 & 65.7 & 50.8 & 64.0 & 48.3 & 63.4 & 64.6 \\
\bottomrule

\end{tabular}
}
\vspace{3mm}
    \caption{Caption \llq{just uncomment for now to see the numbers}}
    \label{tab:low_shot}
\end{table}
\end{comment}

\begin{figure}
    \centering
    \includegraphics[trim={0.9cm 0 0.75cm 0},width=0.97\textwidth]{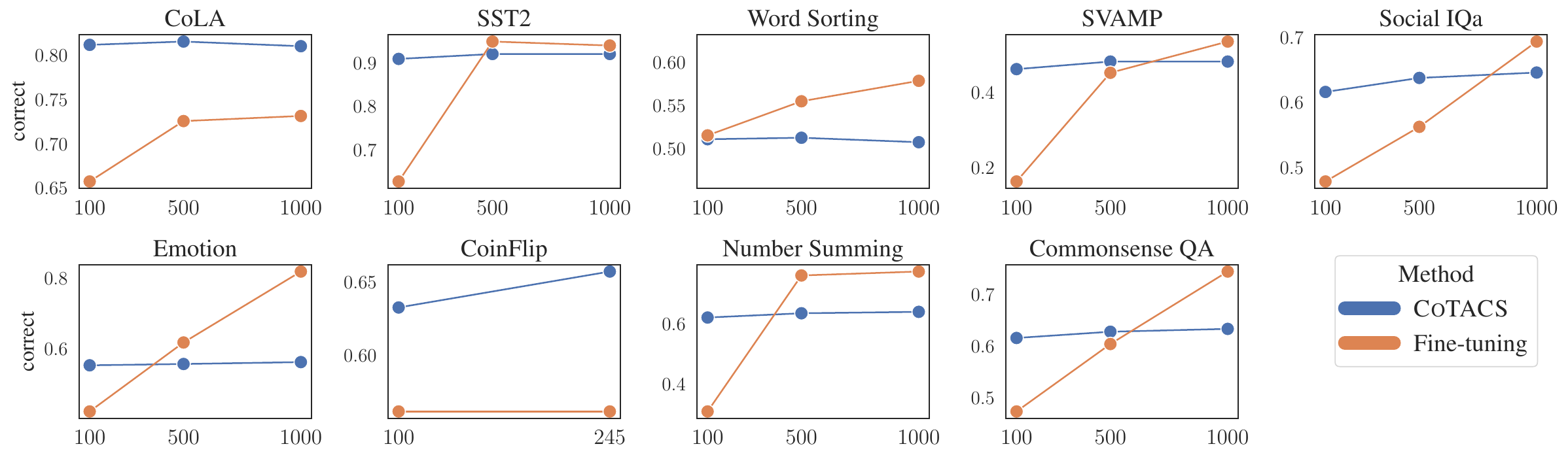}
    \caption{Performance tradeoff between \search{}, which requires saving just a program string, and fine-tuning, which requires saving an entire checkpoint, as we increase the number of training examples. Although fine-tuning typically performs better with more data, \search{} provides an alternative that is lighter-weight and stronger at low-to-medium numbers of instances.
    \vspace{-0.15em}
    }
    \label{fig:finetune}
\end{figure}

 %  and a lightweight alternative to fine-tuning. given larger numbers of examples.

% \section{Discussion}

\subsection{Qualitative Analysis}
\label{sec:qualitative}
Since we rely on the LM as a code emulator, there is no guarantee of correct execution.  The generated intermediate outputs allow us to examine if the model can faithfully emulate the program. We observe failure cases where the LM incorrectly simulates the program execution even if the generated program is correct as shown in Figure~\ref{fig:qualitative_examples_neg}. We also include positive qualitative examples in Appendix~\ref{sec:qualitative_analysis}.

% \llq{do we want some negative examples? not sure if we have enough space though.}
% \nw{I'd say overpopulate the section (so, yes) and then we can cut down when necessary} 

\begin{figure*}
\begin{lstlisting}[frame=single,style=base]
def track_coin_flip(text, num_flip):
    """
    Track the current state of a coin after a certain number of flips.
    Args:
    	text (str): a string containing information about the state of a coin at different stages.
        num_flip (int): the number of flips to consider.
    Returns:
        A dictionary containing (at least) the field 'answer', whose value is of type `str` and contains the state of the coin after the specified number of flips.
        The dictionary also contains the result of the intermediate steps of the reasoning process.
    """
    # Step 1: Identify and extract all mentions of flipping a coin from the input text.
    coin_flip_mentions = extract_coin_flip_mentions(text)

    # Step 2: For each identified instance of flipping, determine whether it is a heads or tails outcome.
    flip_outcomes = {mention: determine_flip_outcome(mention) for mention in coin_flip_mentions}

    # Step 3: Track and record the state of the coin after each flip.
    coin_state = track_coin_state(flip_outcomes, num_flip)

    return {`coin_flip_mentions': coin_flip_mentions, `flip_outcomes': flip_outcomes, `answer': coin_state}
    
Example Input: `A coin is heads up. Gee does not flip the coin. Joseluis flips the coin. Cory does not flip the coin. Stefanie flips the coin. Is the coin still heads up?'

Example Output:  {
    `coin_flip_mentions': [`A coin is heads up', `Joseluis flips the coin', `Stefanie flips the coin'],
    `flip_outcomes': {
        `A coin is heads up': `heads', 
        `Joseluis flips the coin': `@heads@', 
        `Stefanie flips the coin': `heads'
    },
    `answer': `heads'
}
\end{lstlisting}
\caption{Qualitative examples of LLama-2 13B the coin flip tracking task where the model fails to correctly simulate the program and is correct for the wrong reasons.
% \vspace{-0.5em}
}
\label{fig:qualitative_examples_neg}
\end{figure*} 
\section{Related Work}
\vspace{-0.3cm}
\myparagraph{Reasoning via Code.}
Using code for reasoning is a burgeoning area that has shown improved results on many algorithmic tasks \citep{chen2022program, gao-etal-2023-pal}. Many approaches ask LLMs to express their reasoning as code and leverage code interpreters to execute them. Recently, and concurrent with our work, some studies investigate training LLMs as code compilers, where the LM is prompted to emulate code execution \citep{li2023chain, chae2024language, mishra2023prompting}.
These LM-as-compiler approaches fall into a broader category of work that invokes LLMs as subroutines in programs~\citep{kalyanpur2022braid,weir-etal-2024-nellie}.
Different from ours, these works mainly rely on manually prompting very large models, while we focus on training open-source LMs to both generate and emulate programs. 
%\mk{CoC does not do psuedo execution and if we wil cite the other work, we should say it is  is concurrent with ours.}\llq{i thought it does psuedo execution whenever code is not executable?}
%We follow this line of work but use instruction tuning using programs, rather than in-context learning only. 
In addition, we aim to achieve task generalization by searching for an optimal program for a given task---different from \citet{chae2024language} who rely on prompting LMs with specific code instructions. Ours is the first work on code-based reasoning that employs search over the program space with the goal of generalizing an optimal program to a task.

\myparagraph{Prompt Optimization.}
%Optimizing the few-shot prompt used for in-context learning has received a lot of attention.
Our search procedure, \search, has a similar spirit to in-context learning optimization approaches where the goal is to find an optimal set of exemplars (an optimal pseudo-program, in our case) for a given task.
Existing studies \citep{zhang2022active, rubin2022learning, ye2023compositional, gupta2023coverage, khalifa2023exploring} explored various methods to select optimal in-context examples, leveraging similarity- or diversity-based heuristics---to name a few. Searching for useful task instructions has also been explored \citep{honovich2022instruction,promptrank23,chen2023instructzero}.

Another related area of research is automated prompt engineering \citep{shin2020autoprompt, deng2022rlprompt, prasad2023grips} that bootstraps an effective prompt using some reward function. 
While LMs have been shown to be effective at producing their own prompts \citep{zhou2022large, yang2024large, pryzant2023automatic, ye2023prompt}, our work shows that LMs can also reason by generating and executing their own generated programs.
%. \mk{i don't feel good about this last sentence}
% They showed that models can be their own prompt engineers by iteratively generating and improving their own proposed prompts. 
% \citet{honovich2022instruction} also showed that LMs can inductively infer the task description from a few examples.
% DsPY\citep{} proposes a framework for algorithmically optimizing prompts and examplars.
Our method differs from these studies as it uses the same input instruction and optimizes the intermediate representation, rather than modifying it via prompt optimization.
Finding a single program string to solve a class of problems is also related to finding a high-level NL description of a task using one or multiple demonstrations~\citep{weir2023one}.

% instead we optimize the intermediate representation.
% \method asks the LM to produce a prompt via code that can be used to generate the response by invoking a function call, however, \search{} differs by keeping the original prompt the same and only searching over the intermediate programs generalize across the dataset.
% \search{} instead searches over the code space to find programs that generalize across the dataset.

% These programs can then intuitively be used to self-optimize themselves, rather than needed an automated prompt library.

% \paragraph{Instruction Tuning.}
% LMs have been shown to be effective at following instructions when fine-tuned with instructional data \citep{longpre2023flan, sanh2021multitask}. Several instruction-tuning datasets have been proposed \citep{kopf2023openassistant, peng2023instruction, xu2024wizardlm} for fine-tuning. One prominent example of this is Alpaca \citep{taori2023alpaca} which provided a set of instructions. We use their dataset as a comparison to our code instruction dataset.
\vspace{-2mm}
\section{Conclusion}
We present \method{}, a methodology that trains language models to generate and execute their own Pythonic reasoning programs in response to task instructions.
We convert the Alpaca instruction tuning data into \method{} instances that can be used to \method{}-tune any models.
We design an optimization algorithm, \search{}, that applies \method{} models to a new dataset by generating and searching through possible programs that can be reapplied to new task items. Applying the \search{} search algorithm yields task performance that exceeds that of few-shot in-context-learning and typical NL instruction following. 
Our work demonstrates a way to apply LM-based programmatic reasoning to NLP benchmarks that require softer reasoning skills not easily stated in code syntax.

\section{Limitations}
\label{app:limitations}
While our work represents a step towards utilizing code language models for non-algorithmic reasoning tasks, \method still suffers from the following limitations: 
\begin{itemize}[left=15pt,itemsep=2pt] % This removes the indentation
    \item \method is suitable for soft reasoning tasks for which step-by-step programs are difficult to describe. However, when solving algorithmic tasks where a precise step-by-step program is possible, passing the generated code directly to an interpreter may be preferable to emulating code execution via the LM. 
    \item We have found that the LM can occasionally produce a result that is inconsistent with the code emulated, as noted in \autoref{sec:qualitative}. In which case, the code does not faithfully reflect the reasoning process followed by the model. 
    %\item Our finetuning dataset is relatively small and, therefore, may not unlock the full potential of our approach. We hope future work can scale up \method in terms of both model and data sizes. 
    \item There is an extra computational overhead when emulating code execution via an LM compared to using an interpreter as the LM needs to generate intermediate variables along with the final answer.
\end{itemize}

\section{Acknowledgements}
We thank 
Li Zhang,
Valentina Pyatkin,
and Khyathi Chandu 
for feedback on ideas and earlier drafts.
We also thank the organizers of the AI2 Summer 2023 Hackathon during which this project was initially conceived.

\bibliography{anthology-shrunk-1-24,ref}
\bibliographystyle{iclr2021_conference}

\newpage
\appendix

\section{Prompts for Converting Alpaca to \method}
\label{app:prompts}
\autoref{fig:cleaning_prompt}, \autoref{fig:plan_prompt}, and \autoref{fig:instantiator_prompt} display the prompts used to GPT-4 in sequence to convert Alpaca into the \dataset{} dataset. The prompts (1) convert all inputs and outputs into Pythonic types like strings, lists and dicts, (2) generate plans to answer a given instruction, and (3) instantiate each plan as a Python program with underspecified function calls.

\begin{figure*}

\scriptsize
{\tt
\begin{lstlisting}[language=Python]
Clean up the following json items. Map the input and output fields in each item to a proper pythonic item (e.g. list, dictionary, or clean string). It shouldn't have newlines if it's a string. DO NOT INCLUDE "..." in your outputs.

INPUT 1: 
{`instruction': `Classify the following objects by color.', `input': `Ribbon, Tie, Pen', `output': `-Red: Ribbon\n-Blue: Tie\n-Black: Pen'}

OUTPUT 1:
{`instruction': `Classify the following objects by color.', `input': [`Ribbon', `Tie', `Pen'],  `output': {`Ribbon': `Red', `Tie': `Blue', `Pen': `Black'}

INPUT 2:
{`instruction': `Convert the following text into a list.', `input': `The four elements of design are line, color, shape, and texture.', `output': `- Line \n- Color \n- Shape\n- Texture'}

OUTPUT 2:
{`instruction': `Convert the following text into a list.', `input': `The four elements of design are line, color, shape, and texture.', `output': [`line', `color', `shape', `texture']}

INPUT 3: 
{`instruction': `Generate a list of five items a person might need for a camping trip', `input': `', `output': `1. Tent\n2. Sleeping bags\n3. Flashlight\n4. Matches/lighter\n5. Insect repellent}

OUTPUT 3:
{`instruction': `Generate a list of five items a person might need for a camping trip', `input': `', `output': [`tent', `sleeping bags', `flashlight', `matches/lighter', `insect repellent']}

INPUT 4:
{input}

OUTPUT 4:
\end{lstlisting}
}

\caption{Prompt used for converting inputs and outputs of Alpaca items into Pythonic data types.
}
\label{fig:cleaning_prompt}

\end{figure*}

\begin{figure*}

\scriptsize
{\tt
\begin{lstlisting}[language=Python]
Generate a high-level plan with at most 3 steps that a problem-solving artificial agent could use to complete the following problem. If the problem takes in inputs, your plan should be a high-level abstraction that is generally applicable to new inputs, not just the one shown here. 

Instruction: {instruction}
Input: {possible_inputs}

Your output format should be a series of serialized jsons, 1 per line, for each step of the plan.
They should have the format {"number": <step number>", "description": <step description>}
\end{lstlisting}
}

\caption{Prompt used for generating stepwise NL plans for Alpaca items.
}
\label{fig:plan_prompt}

\end{figure*}

\begin{figure*}

% \paragraph{Program instantiator prompt:} 

\scriptsize
{\tt
For the following questions with example inputs and outputs, generate a function that performs the provided high-level steps. The function should return a dictionary with the field "answer": <answer> as well as the values for intermediate decisions. Don't hard code input-specific items whenever possible. You can make external calls to undefined functions as long as the function name describes its purpose.
}
\begin{lstlisting}
Instruction: Generate three antonyms for the word "wonderful".
Input:
Answer: [`horrible', `abysmal', `appalling']
Steps:
1. Search for synonyms of the target word using a thesaurus.
2. Identify antonyms of the synonyms found in step 3.
3. Package the antonyms as the output in the required format.
Code: 
def generate_antonyms(num_words, word):
    """
    Generate antonyms for a given word.

    Args:
    	num_words (int): the number of antonyms to generate for the word
        word (str): a string representing the word for which we need to find the antonyms.

    Returns:
        A dictionary containing the antonyms of the given word, plus the result of the intermediate steps of the reasoning process
    """

    # Step 1: Search for synonyms of the target word using a thesaurus.
    synonyms = thesaurus_lookup(word)

    # Step 2: Identify antonyms of the synonyms found in step 1.
    antonyms_of_synonyms = [lookup_antonyms(synonym) for synonym in synonyms]

    # Step 3: Package the antonyms as the output in the required format.
    all_antonyms = []
    for antonym_list in antonyms_of_synonyms:
        all_antonyms.extend(antonym_list)

    n_antonyms = all_antonyms[:num_words]

    return {
        `synonyms': synonyms,
        `antonyms_of_synonyms': antonyms_of_synonyms,
        `all_antonyms': all_antonyms,
        `answer': n_antonyms
    }

>>> generate_antonyms(3, `wonderful')

Example Output: 
output = {
    `synonyms': [`amazing', `fantastic', `terrific'],
    `antonyms_of_synonyms': [
        [`horrible', `abysmal', `appalling'], 
        [`dull', `disappointing', `unexceptional'], 
        [`awful', `terrible', `dreadful']
    ],
    `all_antonyms': [
        `horrible', `abysmal', `appalling', `dull', `disappointing', 
        `unexceptional', `awful', `terrible', `dreadful'
    ],
    `answer': [`horrible', `abysmal', `appalling']
}

### 

Instruction: Generate ideas for a travel blog for young tourists visiting India

<...> 

### 

Instruction: {instruction}
Input: {input}
Answer: {output}
Steps:
{steps}
Code:

\end{lstlisting}
\caption{Prompt used for instantiating Python programs from NL plans. See repository for full-length prompt.
}
\label{fig:instantiator_prompt}

\end{figure*}

\section{Further Qualitative Analysis}
\label{sec:qualitative_analysis}
\autoref{fig:qualitative_examples_pos} and \autoref{fig:qualitative_examples_pos2} show good qualitative examples generated by \search{}, along with outputs from the 2-shot prompting baseline for the text classification and math reasoning tasks, respectively. We find that \search{} encourages general-purpose code that is generalizable across multiple examples within the same task. It also enables better interpretability by generating outputs of the intermediate reasoning steps.
% \autoref{fig:qualitative_examples_app} depicts one example error made by the \method{} model on the coin flip tracking task. \todo{more comments}

\begin{figure*}

\begin{lstlisting}[language=Python]
def determine_emotion(sentence):
    """
    Determine the emotion expressed in a given sentence.
    Args:
        sentence (str): the sentence for which the emotion is to be determined.
    Returns:
        A dictionary containing (at least) the field `answer', whose value is of type `str' and contains the emotion expressed in the sentence. The dictionary also contains the result of the intermediate steps of the reasoning process.
    """
    # Step 1: Extract all the words from the input sentence and analyze them to understand their context and meaning.
    words = extract_words(sentence)
    word_context = analyze_words_context(words)

    # Step 2: Identify the emotion-related words or phrases in the sentence.
    emotion_related_words = identify_emotion_related_words(words)

    # Step 3: Return the emotion that best fits the context and the emotion-related words identified in the sentence.
    emotion = determine_best_fit_emotion(word_context, emotion_related_words)

    return {`sentence': sentence, `words': words, `word_context': word_context, 
             `emotion_related_words': emotion_related_words, `answer': emotion}

Example Input: i was feeling festive yesterday

Example Output: {
    `sentence': `i was feeling festive yesterday', 
    `words': [`i', `was', `feeling', `festive', `yesterday'],
    `word_context': {
        `i': `first person singular', `was': `past tense', `feeling': `verb', `festive': `adjective'
    },
    `emotion_related_words': [`festive'], 
    `answer': `joy'
}

2-shot Baseline Output: joy
\end{lstlisting}
\caption{Qualitative example of a LLama-2 13B \method{}-generated program for the text classification task on the Emotion benchmark.}
\label{fig:qualitative_examples_pos}
\end{figure*}
\begin{figure*}[ht]
\centering
\begin{lstlisting}[language=Python]
def solve_math_word_problem(word_problem):
    """
    Solve a math word problem.

    Args:
        word_problem (str): a string representing the math word problem.

    Returns:
        A dictionary containing (at least) the field `answer`, whose value is of type `int` and contains the solution to the math word problem.
        The dictionary also contains the result of the intermediate steps of the reasoning process.
    """

    # Step 1: Identify the key numbers and variables from the problem statement.
    key_numbers = identify_key_numbers(word_problem)

    # Step 2: Understand the problem context.
    problem_context = understand_problem_context(word_problem)

    # Step 3: Perform the appropriate mathematical operations to solve the problem.
    solution = perform_math_operations(key_numbers, problem_context)

    return {
        `key_numbers': key_numbers,
        `problem_context': problem_context,
        `answer': solution
    }

Example Input: Paco had 36 cookies. He gave 14 cookies to his friend and ate 10 cookies. How many cookies did Paco have left?

Example Output: {
    `key_numbers': {`initial_quantity': 36, `quantity_given_away': 14, `quantity_eaten': 10}, 
    `problem_context': `Paco had 36 cookies. He gave 14 cookies to his friend and ate 10 cookies.', 
    `answer': 12}
}

2-shot Baseline Output: 20
\end{lstlisting}
\caption{Qualitative example of a LLama-2 13B \method{}-generated program for the math reasoning task on the SVAMP benchmark.}
\label{fig:qualitative_examples_pos2}
\end{figure*}

%\section{Compute Resources}
%\label{app:compute}

\newpage 
% \newpage
\clearpage
\section*{NeurIPS Paper Checklist}

%%% END INSTRUCTIONS %%%

\begin{enumerate}

\item {\bf Claims}
    \item[] Question: Do the main claims made in the abstract and introduction accurately reflect the paper's contributions and scope?
    \item[] Answer: \answerYes{} % Replace by \answerYes{}, \answerNo{}, or \answerNA{}.
    \item[] Justification: The claim of the paper is a novel reasoning paradigm, dataset for finetuning models on the paradigm, and a task-specific optimization algorithm. We show the empirical benefits of the paradigm and algorithm through our experiments.

\item {\bf Limitations}
    \item[] Question: Does the paper discuss the limitations of the work performed by the authors?
    \item[] Answer: \answerYes{} % Replace by \answerYes{}, \answerNo{}, or \answerNA{}.
    \item[] Justification: In \autoref{app:limitations}.

\item {\bf Theory Assumptions and Proofs}
    \item[] Question: For each theoretical result, does the paper provide the full set of assumptions and a complete (and correct) proof?
    \item[] Answer: \answerNA{} % Replace by \answerYes{}, \answerNo{}, or \answerNA{}.
    \item[] Justification: No theoretical proofs are given.

    \item {\bf Experimental Result Reproducibility}
    \item[] Question: Does the paper fully disclose all the information needed to reproduce the main experimental results of the paper to the extent that it affects the main claims and/or conclusions of the paper (regardless of whether the code and data are provided or not)?
    \item[] Answer: \answerYes{}% Replace by \answerYes{}, \answerNo{}, or \answerNA{}.
    \item[] Justification: Hyperparameters and other details are included in \autoref{sec:approach}.

\item {\bf Open access to data and code}
    \item[] Question: Does the paper provide open access to the data and code, with sufficient instructions to faithfully reproduce the main experimental results, as described in supplemental material?
    \item[] Answer: \answerYes{} % Replace by \answerYes{}, \answerNo{}, or \answerNA{}.
    \item[] Justification: Public access to these will be released when not anonymized.

\item {\bf Experimental Setting/Details}
    \item[] Question: Does the paper specify all the training and test details (e.g., data splits, hyperparameters, how they were chosen, type of optimizer, etc.) necessary to understand the results?
    \item[] Answer: \answerYes{} % Replace by \answerYes{}, \answerNo{}, or \answerNA{}.
    \item[] Justification: In \autoref{sec:approach}.

\item {\bf Experiment Statistical Significance}
    \item[] Question: Does the paper report error bars suitably and correctly defined or other appropriate information about the statistical significance of the experiments?
    \item[] Answer: \answerNA{} % Replace by \answerYes{}, \answerNo{}, or \answerNA{}.
    \item[] Justification: We only report single runs, following previous work.

\item {\bf Experiments Compute Resources}
    \item[] Question: For each experiment, does the paper provide sufficient information on the computer resources (type of compute workers, memory, time of execution) needed to reproduce the experiments?
    \item[] Answer: \answerYes{}
    \item[] Justification: In \autoref{sec:exp-setup}
    
\item {\bf Code Of Ethics}
    \item[] Question: Does the research conducted in the paper conform, in every respect, with the NeurIPS Code of Ethics \url{https://neurips.cc/public/EthicsGuidelines}?
    \item[] Answer: \answerYes{} % Replace by \answerYes{}, \answerNo{}, or \answerNA{}.
    \item[] Justification: It follows the code of ethics.

\item {\bf Broader Impacts}
    \item[] Question: Does the paper discuss both potential positive societal impacts and negative societal impacts of the work performed?
    \item[] Answer: \answerYes{} % Replace by \answerYes{}, \answerNo{}, or \answerNA{}.
    \item[] Justification: In \autoref{app:limitations}.
    
\item {\bf Safeguards}
    \item[] Question: Does the paper describe safeguards that have been put in place for responsible release of data or models that have a high risk for misuse (e.g., pretrained language models, image generators, or scraped datasets)?
    \item[] Answer: \answerNA{} % Replace by \answerYes{}, \answerNo{}, or \answerNA{}.
    \item[] Justification: We do not provide any new risks and discusss risks in \autoref{app:limitations}.

\item {\bf Licenses for existing assets}
    \item[] Question: Are the creators or original owners of assets (e.g., code, data, models), used in the paper, properly credited and are the license and terms of use explicitly mentioned and properly respected?
    \item[] Answer: \answerYes{} % Replace by \answerYes{}, \answerNo{}, or \answerNA{}.
    \item[] Justification: They are properly credited via citation and licenses are in their respective Github pages. As they are open-source we do not explicitly discuss the license here and point interested readers to the original authors license.

\item {\bf New Assets}
    \item[] Question: Are new assets introduced in the paper well documented and is the documentation provided alongside the assets?
    \item[] Answer: \answerYes{} % Replace by \answerYes{}, \answerNo{}, or \answerNA{}.
    \item[] Justification: Yes, described in Section 2.

\item {\bf Crowdsourcing and Research with Human Subjects}
    \item[] Question: For crowdsourcing experiments and research with human subjects, does the paper include the full text of instructions given to participants and screenshots, if applicable, as well as details about compensation (if any)? 
    \item[] Answer:  \answerNA{} % Replace by \answerYes{}, \answerNo{}, or \answerNA{}.
    \item[] Justification: No crowdsourcing was done.

\item {\bf Institutional Review Board (IRB) Approvals or Equivalent for Research with Human Subjects}
    \item[] Question: Does the paper describe potential risks incurred by study participants, whether such risks were disclosed to the subjects, and whether Institutional Review Board (IRB) approvals (or an equivalent approval/review based on the requirements of your country or institution) were obtained?
    \item[] Answer:  \answerNA{} % Replace by \answerYes{}, \answerNo{}, or \answerNA{}.
    \item[] Justification:  No crowdsourcing was done.

\end{enumerate}

\end{document}